# Fast and Efficient Skin Detection for Facial Detection

Mohammad Reza Mahmoodi

Department of Electrical and Computer Engineering, Isfahan University of Technology, 8415683111 Isfahan, Iran
Email: mr.mahmoodi@ec.iut.ac.ir

*Abstract*— **In this paper, an efficient skin detection system is proposed. The algorithm is based on a very fast efficient pre-processing step utilizing the concept of ternary conversion in order to identify candidate windows and subsequently, a novel local two-stage diffusion method which has F-score accuracy of 0.5978 on SDD dataset. The pre-processing step has been proven to be useful to boost the speed of the system by eliminating 82% of an image in average. This is obtained by keeping the true positive rate above 98%. In addition, a novel segmentation algorithm is also designed to process candidate windows which is quantitatively and qualitatively proven to be very efficient in term of accuracy. The algorithm has been implemented in FPGA to obtain real-time processing speed. The system is designed fully pipeline and the inherent parallel structure of the algorithm is fully exploited to maximize the performance. The system is implemented on a Spartan-6 LXT45 Xilinx FPGA and it is capable of processing 98 frames of 640*480 24-bit color images per second.**
*Index Terms*— FPGA, Skin Detection, Hardware Implementation, Real-time, Skin Segmentation, Parallel

## I. INTRODUCTION

A very shocking glance at actuarial of sleepy people driving cars reported by National Sleep Foundation's 2005 Sleep in America poll call up for serious reactions. Accordingly, 60% of adult drivers (about 168 million people) have admitted that they have driven a vehicle while feeling drowsy in the past year, and more than one-third (37% or 103 million people) have actually fallen asleep at the wheel [1]! As a matter of fact, more than 1 tenth of those who have nodded off, say they have done so at least once a month and four percentages of them avouched that they have had an accident or near accident since they were too tired to drive. In addition, the National Highway Traffic Safety Administration have estimated that 100,000 police-reported crashes are the direct result of driver fatigue each year resulted in an estimated 1,550 deaths, 71,000 injuries, and $12.5 billion in monetary losses [1]. Fatigue directly affects the driver's reaction response time and drivers are often too tired to realize the level of inattention. A driving monitoring system shall be a promising solution to this dreadful problem. In such systems, several features including different signs and body gestures in the eyes or head such as repetitious yawning, heavy eyes and slow reaction will be exploited as a drowsiness cue. However, an essential step before facial feature extraction in variety of such systems is detection of human skin to diminish the search area for subsequent facial detection. Here, as the head pose or rotation may vary frequently, using skin color cue to detect facial features could be a promising choice. In such applications, both the accuracy and speed of the algorithm is determinative.

On the other hand, the fact that color is one of the most principal features in human face analysis algorithms has been yielded to development of numerous systems that leverage skin segmentation either as a preprocessor, main processor or post processor. In general, assaying face is relatively intensive task by itself. Thus, it would be effective to use a preprocessor in order to reduce the size of images (regions of interests) and sometimes provide real-time operation feasible. One problem with most skin detection methods is their inability in dealing with factors involved in degrading performance of detectors [2].

Accordingly, there is an immediate need to design a real-time, fast and reliable system to segment skin pixels of an arbitrary image. The proposed system addresses both of these problems as a novel skin detector is proposed and it is implemented in the FPGA. Operating up to 98 frames per second, the whole system successfully detects skin pixels in real-time.

The rest of the paper is as follows. In section 2, previous works in the literature are briefly explored. In section 3, the proposed algorithm is comprehensively presented. Experimental results are illustrated in section 4 and finally, conclusion is provided.

## II. PREVIOUS WORKS

Explicitly defined methods are based on a set of rules derived from the skin locus in 2D or 3D color spaces. These methods have a pixel-based processing scheme in which for any given pixel, rules are investigated to decide on the class of that pixel. Kovac et al. [3] proposed a method of explicitly defined boundary model using RGB color space in two of daylight and flashlight conditions which has been reutilized in [4].

There is no explicit definition of probability density function in non-parametric statistical techniques. Single histogram based Look-UP-Table (LUT) model is a common approach in which the distribution of skin pixels in a particular color space is obtained using a set of training skin pixels. Considering RGB as a color space with finest possible resolution (i.e. 256 bins per each channel), the 3D RGB histogram (or LUT) is constructed from 256*256*256 cells each representing skin probability of one possible $R_iG_iB_i$ value. The learning process is though simple, populating the histogram requires massive skin dataset plus huge storage space. This huge storage requirement is histogram model's blemish which is addressed by using both coarser bins and

2D color models. Jones et al. [5] employed this technique for person detection. In contrary to above LUT approach, the Bayesian classifier considers two histograms of skin and non-skin pixels due to the overlap between skin and non-skin pixels in different color spaces (Jones et al. [5] in their study showed, 97.2% of colors which occurred as skin also occurred as non-skin). Parametric statistical models such as single Gaussian models (SGMs), Gaussian mixture models (GMMs), cluster of Gaussian models (CGMs), Elliptical models (EMs), etc are developed to compensate LUT shortcomings. In addition, they generalize very well with a relatively smaller amount of training set.

In skin detection, ANNs (Artificial Neural Networks) have been utilized for illumination compensation, dynamic models, in combination with other techniques and direct classification. Al-Mohair et al. [6] employed simple 3-layer MLP classifier using different color spaces and different number of neurons in hidden layer. Among common color spaces, they concluded that YIQ gives the highest skin and non-skin separability.

Using online information of the image or sequence of frames has been exploited as an effective idea to counteract non-uniform illumination to some extent. Adaptive models are developed in an effort to present models which are calibrated to given inputs. Yang et al. [7] integrated an ANN system to dynamically configure a Gaussian model. Firstly, according to statistics of skin color pixels, the covariance and the mean value of Cb and Cr with respect to Y channel is calculated, and then it is used to train a neural network which gives a self-adaptive skin color model based on an online tunable Gaussian classifier. Ibrahim et al. [8] employed a face detector to segment face regions. Based on face information, the exact boundary of an explicit method in $YC_bC_r$ is obtained. Using this dynamic threshold, the skin regions all over the image is obtained.

The strategy in spatial-based techniques is extraction of initial skin seed by means of a precision oriented pixel based detector and annex other skin pixels to the seed [9]. Ruiz-de-solar et al. [10] proposed a method of controlled diffusion. In seed generation stage, they utilized a GMM with 16 kernels in $YC_bC_r$ color space and then the final decision about the pixel's class is taken using a spatial diffusion process. In this process, if Euclidean distance between a given pixel and a direct diffusion neighbor (already skin pixel) is smaller than a threshold value, then the propagation occurs. Also, the extension of the diffusion process is controlled using another threshold value, which defines the minimal probability or membership degree allowed for a skin pixel. This process works well in regions where the boundary of skin and non-skin pixels are sharp enough unless the leakage occurs. The effectiveness of this method is questionable as in most conditions, being near a skin pixel is not a concrete clue to be used. Also, using GMM with any number of kernels and any threshold is not reliable for finding seed points.

In our previous works [9,11], we proposed new algorithms in which boosted the perfromance of skin detection considerably. We compared the performance of our system with other methods using an standard dataset [12].

In our recent paper [11], skin pixel in initial seed was extracted using a simple algorithm. Afterwards, each skin pixel starts to "propagate" to other non-identified pixel on virtual lines. In parallel, the edge map of the image is calculated using an empirical threshold to eliminate leakage effect in former methods. The feature used in diffusion was based on the concept of Otsu multi-thresholding. Otsu segmentation was used in several color channels in order to specify the homogonous regions of an image. For diffusion of a master pixel into an under test pixel, several factors are considered including the group in the initial seed (white, gray or black) and the homogenous class of both pixels. In more recent method [11], not only several features were appended including motion information in local regions, color distance map and skin probability map to the former works, but also an efficient method of conservative extraction of initial seed and multi-step diffusion were used.

The result of the two-stage diffusion method was promising but the shortcomming was the speed as the real-time implementation of the system using a modern CPU/FPGA was off-the-table. Though ASIC implementation was a solution, we decided to choose a low-cost solution. In the proposed algorithm, a significant of the image is elminiated using a very fast and effective method. The candidate windows are extracted by projecting the image into a ternary system and using a three-step segmentation. Then, a high perfromance classifier is used. As it will be shown, without dimishing the fidelity of our system signifcantly in favor of speed and cost, the proposed system detects skin pixels effectively. This skin detector as it will be shown can be used for many applications including face and fatigue detection.

### III. PROPOSED ALGORITHM

The proposed skin detection method is designed to be a pre-processing stage for a couple of advanced methods such as Viola-Jones, deep neural networks and many other window based approaches. The goal is to find all the pixels that contain human skin in a given image, while eliminating all other regions as much as possible. The given image is processed through several stages and finally the candidate windows are reported. Subsequently, the skin detector core is performed locally to detect final results.

#### 1. EFFICIENT PRE-PROCESSING

The optimum performance is not dependant on the employed color space [2]. Nevertheless, in this paper, the input image is converted into $YC_bC_r$ color space for some reasons. First, compared with RGB, redundancy of its channels is much lower and its components are highly independent. Secondly, it is widely used for compression works, e.g., in European television studios, JPEG, MPEG, CCD camera and it makes it easy to use it in many applications without a need to transform the image into another color space. Thirdly, as it has been statistically analysed in [13], the performance of $YC_bC_r$ is superior compared to others when using the same method for different color spaces. Also, one advantage of $YC_bC_r$ to many other color spaces like HSV and TSL is that it can be obtained via a linear transformation from RGB format. $YC_bC_r$ is similar to the human visual perception and it is a discrete color space with tight cluster that makes it easy to realize clustering algorithms and work with it [14].

The proposed skin detection method includes a training stage which is carried out offline. Through this stage, necessary statistical data are obtained. These data are used next to partition the color space into 3 different sets. These sets are used to build a ternary image subsequently in the pixel based stage. For every input image, each of its pixels is processed in a pixel-based, neighbour-based and a window-based process. In order to train the system, a standard skin detection dataset [12] was used. The intensity heat maps of the pixels are depicted in Figures 1, 2 and 3.

Based on these gathered data, three sets of pixels are defined. These sets are calculated in accordance with intensity heat maps. Each map is divided into 3 non-overlapping partitions. The boundary of each sector is estimated by a polygon in order to preserve the simplicity of the algorithm and boosting the system in the inference mode. For each map, 2 polygons are estimated. The inner polygon represents the boundary between the pixels which have been observed with high frequency and the pixels which are not highly repeated, but they can still be part of skin in some conditions. The second boundary separates pixels which have been observed with low frequency and pixels which are not observed ever. This has to be carried out for each of the three maps. Based on above definitions, set $T_1$ contains pixels that are simultaneously located inside all three inner polygons. $T_2$ contains pixels which are not included in $T_1$ and they are inside the outer polygons. And set $T_3$ is filled with the rest of pixels. By this partitioning, dealing with nonlinear illuminations will be much easier. The training stage provides coefficients of the polygon lines which are kept in memory of the detection module.

### A. Pixel-based Segmentation

Segmentation of an arbitrary image into a ternary one is the first step. Here, contrary to other methods, the image is not classified into a skin and non skin pixels. Based on so called sets, the image is segmented to 3 groups of pixels. Thus, there are lots of uncertain pixels due to illumination and imaging device effects and other reasons. Making decision on a pixel that whether or not it belongs to a skin is difficult. Therefore, instead of roughly discard or accept it, it will be put aside for further processing steps based on its neighbour and region information.

In pixel-based processing, each pixel of the input image is classified according to above three defined sets. $T_1$ pixels are set white (255), $T_3$ pixels are set black (0) and $T_2$ pixels are set gray (128). By this, a ternary image is obtained. Fig. 4 shows an image selected from SDD [12] along with its ternary conversion. The figure shows that the logic behind the idea of ternary image is reasonable. Some skin regions are detected easily as they are not very affected by non linear illumination, and they more or less will be specified by most methods (they are located in $T_1$ set and shown by white points). Gray pixels are the ones that no strict decision can be made for them. Some of them are truly skin pixels, and some are not.

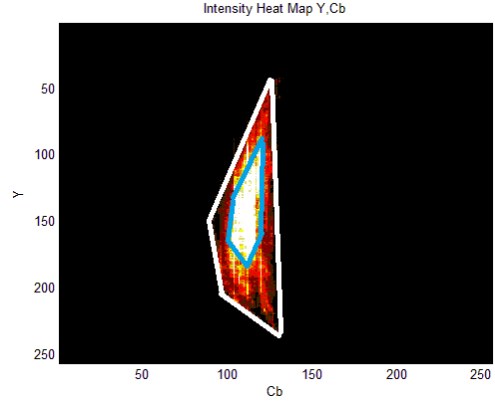

Figure 1. Intensity Heat Map Y,Cb

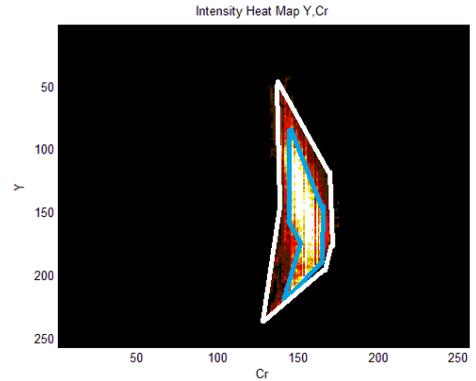

Figure 2. Intensity Heat Map Y,Cr

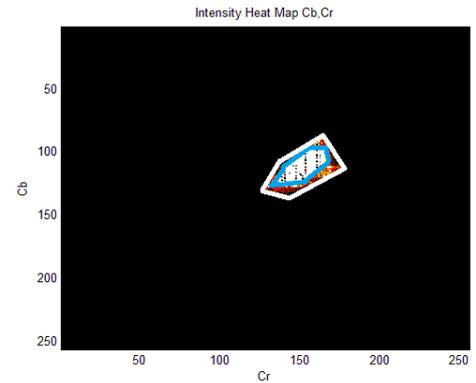

Figure 3. Intensity Heat Map Cb,Cr

### B. Neighbour-based Segmentation

Except for bordering pixels, and skin pixels which have been verified before ($T_1$ pixels), each pixel of the ternary image is processed in a neighbour based algorithm. To that end, 3*3 and 5*5 neighbourhood are considered for each pixel and a score is calculated based on the number of $T_1$, $T_2$ and $T_3$ type pixels in its neighbours, and by using a set of rules on the number of pixels in neighbours. The score is calculated by (1), in which ζ is the total score, K is a constant, T is the score in the 3*3 neighbour and ϕ is the same in the 5*5 neighbour.

$$\xi = K \times T + \Phi \qquad (1)$$

Decision is made based on the value of ζ. Two thresholds have been statistically determined in order to decide whether a pixel is skin or not. Thresholds have been obtained empirically by using the classifier in the inference mode and through a backpropagation optimization procedure. A pixel with score lower than $th_1$ is considered black and higher than $th_2$ is considered white. The output of this stage is another ternary image that for some of its pixels, decision has not been made yet (skin or not skin). It has been observed that for many skin pixels with considerable number of skin pixels around them, they have not been set white yet, and this may be due to illumination condition (Fig. 4), a shadow, moustache, etc. The proposed neighbor-based procedure compensates these effects. Also, isolated pixels mainly caused by noise will be eliminated.

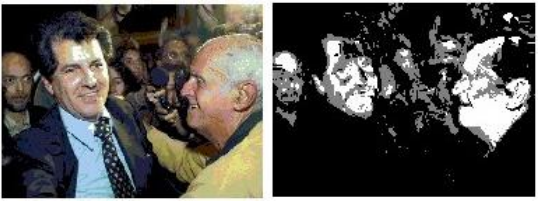

Figure 4. A sample image and its ternary conversion

### C. Window-based Segmentation

Without hurting the generality of the problem, the size of output windows is chosen to be 16×16. Each window is to be determined as candidate or not. The procedure is simple. If there is no white point in the window, then it is reasonable to bypass it, and if it contains a considerable number of white points, there is a high probability that this window includes a whole or at least part of human skin. Thus, a set of rules are utilized in order to filter many of windows. There is a minimum number (threshold) for both white and gray points which determine the validity of a window. An important feature considered in setting thresholds and rules for different parts is that the algorithm should never miss any window (as much as possible) and false positives will be filtered by other post advanced methods.

To improve the results, after window processing, a morphological operation is performed to add parts of a face which probably are occluded by a glass or even eyes (They have colors different from skin color). Therefore, windows which are surrounded by candidate windows will be annexed to those previously chosen. Fig. 5-6 represents a couple of images selected from our database and the result of applying the algorithm on them.

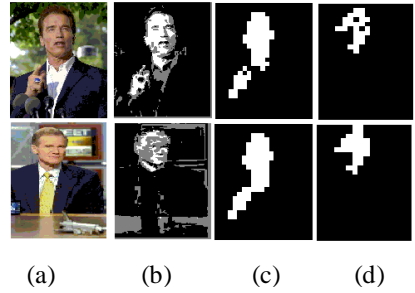

(a)     (b)     (c)     (d)

Figure 5. Example images with linear illumination, a) input image. b) Ternary image. c) candidate windows d) filtered windows

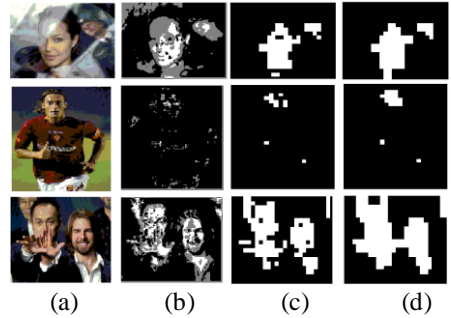

(a)     (b)     (c)     (d)

Figure 6. Example images with nonlinear illumination, a) input image. b) Ternary image. c) candidate windows d) filtered windows

### D. Performance assessment of pre-processing step

Besides using common classification metrics such as FP, and TP, another criterion "ER" is defined here to evaluate the performance of the pre-processing algorithm. Elimination rate (ER) is the ratio between the number of output windows and the number of windows in an image. It is clear that this is disparate for different images as one image can contain a couple of faces and also in some condition, there are non-face regions like hands reported to be face. ER represents how effectively this pre-processing algorithm reduces non-skin parts of an image, and how combination of this algorithm and other algorithms can boost both the speed and accuracy of the final system. The pre-processing has been tested on other datasets, LFW [15], Pratheepan [16] and Bao[17]. In Fig. 7, the results are depicted. Fig. 8 shows ER rate for images of LFW data bases.

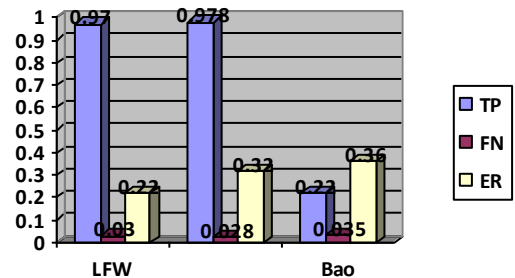

Figure 7. Performance of pre-processing window identifier in different datasets

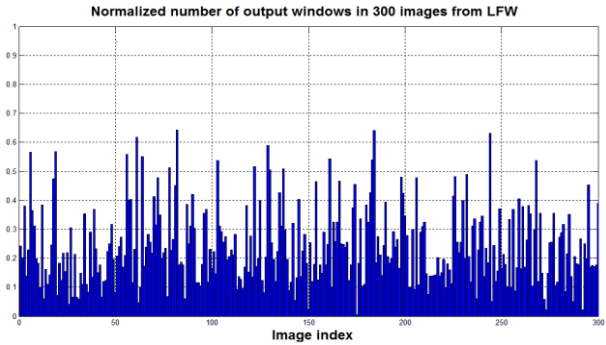

Figure 8. Normalized number of identified windows for 300 images in LFW (TP=%97)

Two important points should be mentioned. First of all, it is possible to decrease TN rate more by changing the threshold values (simply there is a trade-off between ER rate and TN rate). Considering a face detection algorithm, average ER rate of 22% is very promising to reduce the computation complexity and besides that, it should be noted that TN rate of 3% doesn't directly affect the performance of the face detector. In order to clarify this point, two examples where the pre-processing hasn't performed well is presented in Fig. 9. In the upper image, the algorithm though has detected all skin regions, couldn't deal with background skin-like issue and ER is not appealing (although it is still good to eliminate 50% of the image). The second example shows the case where the illumination is extremely nonlinear (the color of the man's face is rather reddish) and hence the algorithm fails to detect one skin region completely and one incomplete. By applying these image to a typical MLP-based face detector, both faces in the upper image are detected and for the second image, only the right face is detected. Notably, by increasing the thresholds of the algorithm, it is possible to reach the point where both of the faces could be detected, but that comes with higher ER rate.

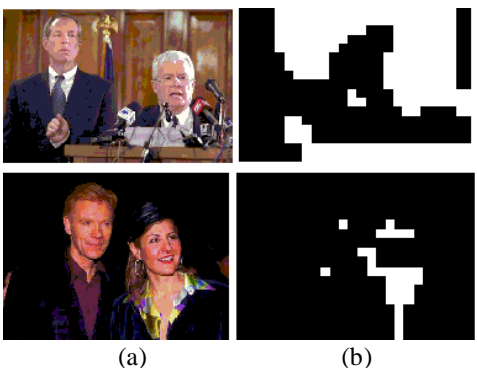

(a)　　　　　　　　(b)

Figure 9. Example images with poor pre-processing: a) input images, b) identified candidates

### 2. SKIN SEGMENTATION CORE

In the following, a reliable skin detection algorithm is applied to those pre-identified windows locally to generate final results. This not only increases the speed of the system, but also simplifies FPGA implementation which is the main goal. Besides that, simulation result shows that the classification accuracy does not drop significantly.

Here, traditional idea of spatial analysis i.e. generation of seed and propagation based on each of candidate point is employed. However, method of seed extraction, diffusion procedure, and features which are used in propagation process are different from previous works. In addition, controlled feedback is also derived from previous frames to improve the performance. The data flow of processing each identified window is depicted in Fig. 10.

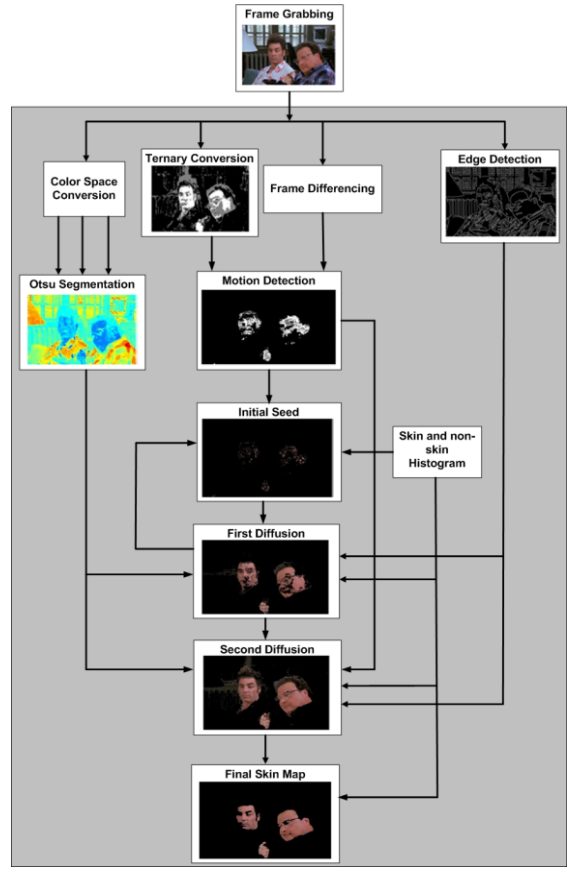

Figure 10. Main Skin Detector Core

The ternary conversion has been already obtained for the whole image. In parallel with that, color space conversion, and frame differencing and Otsu segmentation was also applied on these channels in order to specify the homogenous regions of an image. This is the fundamental calculation of the homogenous regions. Also, a simple edge detector is used to detect edge points. As it will be shown, the result will be employed in both diffusion steps to avoid leakage as much as possible.

The process starts with acquisition of one identified window from the pre-processing stage. For all windows, skin pixels which have "moved" are detected. The result will be utilized in accompanying with Bayesian classifier to generate initial seed. Note that hardware implementation of all of these blocks can be simple and fast by exploiting the parallel nature of the FPGA as well as appropriate memory management.

Skin points in initial seed are those from which propagation start. The first diffusion is designed conservatively to add other candidate pixels from which diffusion may start. First diffusion is only based on

"homogenous" feature as it will be elucidated in next sections. Second diffusion is however, based on several additional features such that most other non-identified skin pixels are annexed. The final skin map is determined as a result of filtering and rectifying the second diffusion output. In the following, each step of the algorithm is illustrated in detail by the aid of graphical representation of performing different parts of the algorithm on general images and videos.

### A. Motion Detection

Motion is a very important feature in video images and it has been leveraged in the proposed paper with two aims. Firstly, it is used to help construction of seeds, and secondly, it is exploited as a feature in diffusion. One conventional and simple approach is frame differencing technique. In this case, consecutive frames are differentiated and thresholded to generate a motion mask. However, there are tangible problems which may affect the performance seriously. One is the case of moving cameras in which fixed objects (pixels) may seem ambulant, and the other is the case of movement of non-skin objects. This is the reason of combining ternary image with the result of frame differencing. In this case, most uninterested pixels which are falsely detected due to the aforementioned reasons will be filtered out. Fig. 11 shows some sample results of this approach. Left images represent ternary conversion of several color images, the middle one are the result of using only frame differencing, and right images are outputs of using proposed approach. This skin-motion detection scheme is robust in generating required points in skin regions and as a result, it has been directly utilized in generating initial seed. Here, it is not assumed that all skin pixels are detected. Practically, at least certain number of pixels in each skin patch should be extracted. Throughout of this paper, a pixel is called "ambulant pixel", if it is selected in motion detector (right image).

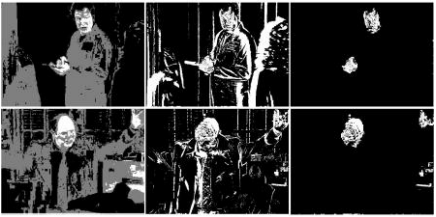

Figure 11. The effect of using ternary image in motion detector

### B. Initial Seed Generation

Determining root pixels or initial seed is extremely decisive in final mask as it can lead to both high false detection and rejection. Consequently, the method conservatively selects pixels. Most former spatial based methods have used Bayesian classifier or GMM with a relatively high threshold to detect skin seed. However, practically, these methods are not efficient. The main reason is that Bayesian classifier alone, even with a huge training set, is not capable of detecting skin pixels in many ordinary conditions without false alarms. Bayesian performance is extremely dependent on the training images. In addition, false alarms in former spatial based methods were too influential on the final result and no particular method was considered to control that. The performance of other traditional methods such as MLP based modes, parametric models, explicitly defined boundary methods, etc has been reported to be even worse compared to Bayesian [5].

In order to address problems of these methods, motion detector, a feedback mechanism from first diffusion of former frame and a modified Bayesian classifier are fused. The initial seed mainly depends on the score of a pixel in both skin histogram and non-skin histogram and also on whether or not it is an ambulant pixel. Jones et al. [5] introduced the application of concept of Bayesian theory on skin classification. According to Bayes rule, Eq. 1 is rewritten as:

$$P(skin|R_iG_iB_i) = \frac{P(|R_iG_iB_i| \, skin)P(skin)}{P(|R_iG_iB_i| \, skin)P(skin) + P(|R_iG_iB_i| \, non-skin)P(non-skin)} \quad (2)$$

Using each of skin and non-skin histograms, $P(R_iG_iB_i|skin)$ and $P(R_iG_iB_i|non\text{-}skin)$ will be calculated respectively. In designing Bayesian classifier, two assumptions are possible for prior probabilities. However, it is possible to use following detection rule to eliminate the effect of priori probabilities:

$$\frac{P(R_iG_iB_i|\,skin)}{P(R_iG_iB_i|\,nonskin)} > \theta, \; \theta = K(\frac{P(non-skin)}{P(skin)}) \quad (3)$$

where K is a tunable parameter to remove the dependency of detector's behavior to priori probabilities. The optimum value for θ is obtained using ROC curve depend on the application. Jones et al. [5] mathematically proved that for any choice of priori probabilities, the resultant ROC is the same. In practice, there are several problems associated with Bayesian classifier. Generally, nonparametric methods do not generalize well and require extremely huge number of training pixel to produce relatively acceptable results. In fact, both histograms of skin and non-skin pixels should be huge enough to classify well. For a 24 bit RGB image, there are $256^3$ possible colors for each pixel. When using Eq. 3 for making decision about a pixel, two problems may occur (and they frequently happen). First of all, if the denominator i.e. $P(R_iG_iB_i|non\text{-}skin)$ is a small value (as a reason of inadequate number of that specific color in non-skin training set), the ratio of probability will inaccurately go up. In this case, although the numerator may have been small, the pixel will be probably detected as a skin. The second problem originates from a similar point, but in reverse order. If a pixel has been frequently found in skin images and it has been observed in several non-skin images with pseudo skin color backgrounds, then the ratio would be unrealistic and unreliable again

To manage Bayesian shortcomings to generate initial seed, in addition to the use of ambulant pixels in generation of seed which addresses the second problem to some extent, not only the ratio of probability of a pixel to be skin and non-skin is utilized, but also the pure probability of a pixel to be skin is used. To be a skin pixel, the pure probability should be higher than a particular threshold. This compensates the first side effect of using Bayesian classifier alone (if the denominator i.e. $P(R_iG_iB_i|non\text{-}skin)$ in Bayesian rule is a small value (as a reason of inadequate number of that specific color in non-skin training set)). Another feature which is utilized in the proposed method is ternary based motion image which boosts the performance, reliability and robustness to some extent. Those points which are not ambulant must achieve a high score in Bayesian classifier to be considered when populating initial seed. Also, the first diffusion is designed conservatively and experimentally it has been observed that most points extracted after first diffusion are truly skin pixels. Thus, after processing any candidate window, the result of first diffusion should be preserved to be used in seed generation of next frame. Selected points in first diffusion of each frame will be candidate to be in the seed set of the next frame with lower threshold. In Fig. 12, it is shown that in each skin patch, only several skin pixels are detected, and in subsequent stages, they will be employed for diffusion. This shows the conservativeness feature of the proposed seed generation approach.

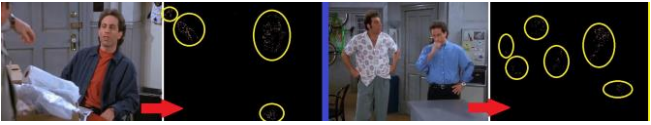

Figure 12. Initial seed extraction for two images

*C. Homogeneous Region Extraction*

Otsu method of image segmentation is a nonparametric and unsupervised method of automatic selection of an optimal threshold. The procedure is very simple, utilizing only the zeros and the first order cumulative moments of the histogram. Using Otsu on specific color channel, the image is segmented to a predefined number of classes (multi-thresholding segmentation). It has been observed that in most of the images, there are certain number of pixels which are related to the human skin but have been excluded in the initial segmentation. Fortunately, in chromatic color components, these pixels are grouped in the same class or semi-class of skin pixels in the seed (homogeneous regions). This feature of many initially unidentified skin pixels is leveraged in both two diffusions in order to annex many skin pixels to their actual group (skin). Approximate homogenous regions are estimated using fusion of Otsu segmentation in several 4 color channels. Fig. 13 represents the "RGB exhibition" of performing multi-thresholding Otsu segmentation on I color channel of YIQ color space for several images. The figure shows that the idea is clearly justifiable. Several color components are fused, so the result would be more robust in a general imaging condition. Comparing all features used in diffusion stage, the homogeneity feature which is derived through Otsu segmentation is the most positively influential factor.

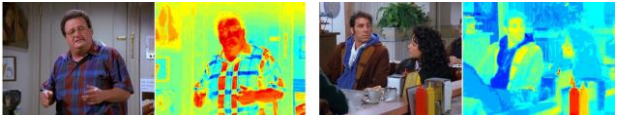

Figure 13. Otsu segmentation on I color channel for several images

*D. Diffusion*

The idea of diffusion is propagation from certain skin pixels to other non-initially-specified skin regions. In both diffusions, for each white point candidate, the process starts with nearest neighbor pixels and continues to surrounding points. One difference between two diffusion stages is that in the first diffusion stage many other probable skin pixels are added to the initial seed and these pixels themselves can be leveraged for finding other skin pixels in the second diffusion stage (a new seed is constructed for second diffusion). Though, in the second diffusion stage, the positive diffused pixels will not be utilized for new propagation again.

The first diffusion is only based on homogeneous feature which is extracted by means of Otsu segmentation. However, the conservativeness of the algorithms is still preserved to avoid extension of false alarms in subsequent step. Thus, each under test pixel is tested to check if it is an ambulant pixel or not. Being an ambulant pixel does not guarantee that a pixel to be located in initial seed; as there are many ambulant pixels which are not necessarily skin pixel. If a pixel has been an ambulant one, it imposes a new condition which should be investigated. Based on Otsu segmentation in different color channels, the value of pixel is thresholded using a weak threshold for ambulant pixels and a stronger threshold for others.

Second diffusion is quite different from the first one. Even though it obeys the similar approach for propagation, several features are employed to determine diffusivity. As a matter of fact, for each particular under test pixel, a diffusion function is calculated as:

$$F(x) = \sum_{i=1}^{5} w_i \, f_i(x) \quad (4)$$

Where $f_i(x)$ is the feature score of under-test pixel "x", $w_i$ indicates the weight of each feature score, and $F(x)$ equals the total score of the diffusion which will be harshly thresholded for final decision making. Features are homogeneity, distance, probability, motion and a feedback from previous frame. The concept of homogeneity was already illustrated in previous sections, here the score i.e. $f_1(x)$ is calculated as:

$$f_1(x) = \exp(-\alpha \sum_i d_i) + \beta \quad (5)$$

where $\alpha$ and $\beta$ are scaling factors and $d_i$ is related to the difference between uniformity class of master pixel and under test pixel in different color channels. Distance feature has been already employed in [19] for an spatial analysis based skin detection. However, using it alone is not useful in many conditions since there are many non-skin pixels around skin objects especially in boundary regions. In proposed approach, perching near a skin pixel is considered a very important cue to determine skinness. Probability also does not strictly determine the skinness, but its value may be utilized in order to strengthen the decision power.

The output of the second diffusion with all of the aforementioned arrangements still needs a filtering and refining process to be acceptable. In this case, a relatively low threshold Bayesian classifier filters probable non-skin points particularly near boundary regions or pixels which have been diffused due to the leakage. Consequently, final mask will be obtained (Fig. 14). As the figure shows, the proposed method is capable of dealing with wide variety of imaging conditions, disparate illumination conditions and complex backgrounds.

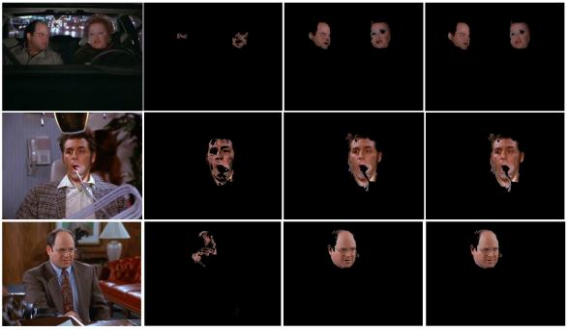

Figure 14. Diffusion steps in proposed algorithm (first diffusion, second diffusion, and final mask)

To avoid the leakage in both diffusions, edge map of the image is thresholded to eliminate many unnecessary details and preserve strong edges to stop diffusion. In the first diffusion step, most edge points are preserved to support the idea of conservativeness propagation. But in the second diffusion step, most of the relatively weak edges are eliminated to allow apposite propagation (Fig. 15). Red points in the middle images demonstrate how much the propagation has been penetrated through the edge lines after filtering (final mask).

### E. Discussion

In this section, we introduced a skin segmentation algorithm which effectively classify each pixel of an arbitrary frame of a video into skin and non-skin classes by using a novel two-stage diffusion method locally on pre-filtered windows. The system is very performance efficient. Quantitate and qualitative results both support the idea that using this detector in face detection and fatigue detection systems [35,36] could be very promising. The algorithm is very speed efficient though it needs pre-processing, most computational steps can be implemented in parallel which makes the system very fast. Skin Detection Hardware

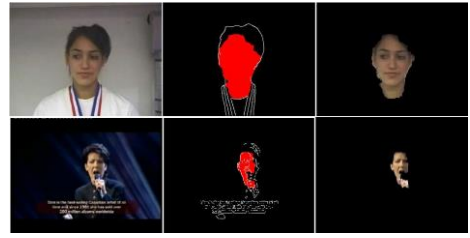

Figure 15. Two sample images, corresponding edge map and final skin

### F. Hadrware Implementation

The system has been implemented using the image processing framework which introduced in [18]. The skin detection core is depicted in Fig. 16. The main blocks include the pre-processing unit and 8 skin core modules. For the pre-processing block, basic blocks are designed in parallel using high performance FPGA fabric DSP resources. 52 bits are considered for each pixel in which 24 bit includes the main and the rest is the tag for each processing unit. For example, one bit for detecting motion. Due to the inherent parallelism in the algorithm, a significant speedup in compare with software implementation is gained. Incoming pixels are preprocessed by incoming data rate (125MHz (byte rate)), and following computation of the motion and edge tags by means of frame differencing and simple filtering, the result is sent to the buffer FIFO to be stored in the external SDRAM.

Control unit at the top level architecture manages the 8 skin detector cores. Each of them has their own control block. Each window is fetched from the MCB (Memory Control Block), after a simple window processing which is the last stage of pre-processing, the window (8*8) will be kept for further processing or the next window will be fetched. The stride is 4 pixels in both directions. The result after each operation is executed and it will be written in its related tag. Finally, the skin mask is obtained.

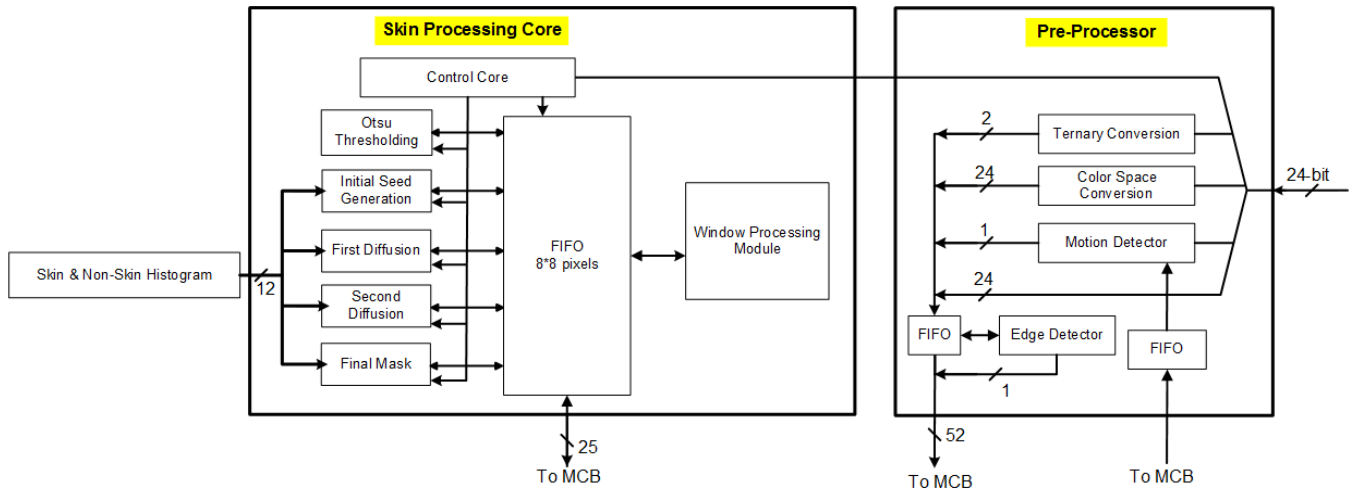

Figure. 16: Pre-processing and Skin Core Module

## IV. EXPERIMENTAL RESULTS

In this section, quantitative results are presented. Unfortunately, no standard dataset suitable for video skin segmentation algorithms is compiled yet. The only one is Feeval [19], which consists of 8991 frames embedded in 25 videos with manually annotated ground truths. However, the videos lack enough quality in most cases and also, the ground truths are not precise. To provide quantitative results, the performance of the proposed method is compared with previous works by using a self-made database which consists of 33 videos and ground truth has been labeled for 106 random frames. Most videos include challenging scenes. In ground truth (GT) annotation, pixels are classified into skin, non-skin and undecided pixels and this has been the key factor for production of accurate GTs. Undecided pixels are mainly those pixels located in skin and non-skin boundaries. In order to present a fair comparison, several algorithms which have exploited skin detection for the task of face detection were tested on the provided database.

Quantitative results are promising. Table. I show the performance of the proposed method in compare with several former approaches using several standard criterions defined in [5]. Results are based on applying each algorithm on all images with GTs.

TABLE I.    THE PERFORMANCE OF DIFFERENT METHODS

| Method | Precision | Recall | F-score |
| --- | --- | --- | --- |
| Vadakkepat el al. [21] | 0.5411 | 0.4046 | 0.4630 |
| Hsu et al. [22] | 0.4996 | 0.6796 | 0.5758 |
| Albiol et al. [23] | 0.4004 | 0.7331 | 0.5178 |
| Sagheer et al. [24] | 0.4470 | 0.2894 | 0.3514 |
| Li et al. [25] | 0.2809 | 0.7313 | 0.4058 |
| Huang et al. [26] | 0.3688 | 0.7166 | 0.4868 |
| Pai et al. [27] | 0.1834 | 0.2866 | 0.2236 |
| Wu et al. [28] | 0.4920 | 0.6286 | 0.5518 |
| Thakur et al. [29] | 0.4526 | 0.6879 | 0.5460 |
| Yutong et al. [30] | 0.4726 | 0.7121 | 0.5681 |
| Qiang-rong et al. [31] | 0.5508 | 0.5634 | 0.5570 |
| Subban et al. [32] | 0.2986 | 0.5509 | 0.3872 |
| Chen et al. [33] | 0.2936 | 0.7080 | 0.4150 |
| Anghelescu et al. [34] | 0.1366 | 0.7331 | 0.2302 |
| Wang et al. [35] | 0.5748 | 0.5465 | 0.5603 |
| Proposed | 0.5830 | 0.6135 | 0.5978 |

The experimental setup is depicted in Fig. 17. The camera sends 24-bit color images to the FPGA via provided Ethernet connection. After executing initial necessary connection commands, camera sends an ARP request to the FPGA and then, video frames are transmitted to the FPGA sequentially after MCB calibration is finished. Xilinx Integrated Software Environment, ISim and ChipScope analyzer were used for implementation, simulation and debugging purposes respectively. The system has been implemented using a Spartan-6 XC6SLX45T-FGG484 -3C FPGA and an off-chip 128 MB DDR3 memory.

The result of the proposed skin segmentation algorithm for is depicted in Fig. 17. In this figures, the live video which comes from the camera is shown in the left monitor and the processing result is shown on the right one. Practical implementation of the system is verified on 30 fps VGA 24-bit images; however, simulation result shows that the system can process up to 98 fps.

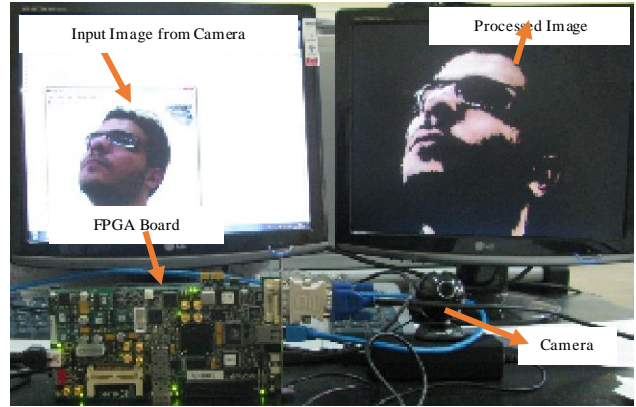

Figure 17. Experimental Setup

## V. CONCLUSION

A skin segmentation algorithm was proposed in this paper based on a very fast efficient pre-processing and a local two-stage diffusion method which has a F-score accuracy of 0.5978 on SDD dataset and it outperforms other methods in this same dataset. The system was implemented using an FPGA to achieve real-time specifications. The hardware system includes a data acquisition block, a processing unit and a DVI interface block and dynamic data transmission between the Ethernet stack, memory and the skin detection unit makes the whole system work real-time. Both of the stack and detection core are designed fully pipeline and inherent parallel structure of the algorithm is fully exploited to maximize the performance. The system is implemented on a Spartan-6 LXT45 and it is capable of processing 98 frames of 640*480 24-bit color images per second.